\documentclass[10pt,twocolumn,letterpaper]{article}

\usepackage{wacv}      % To produce the REVIEW version for the algorithms track
\usepackage[accsupp]{axessibility}
\usepackage{lipsum}

\definecolor{wacvblue}{rgb}{0.21,0.49,0.74}
\usepackage[pagebackref,breaklinks,colorlinks,allcolors=wacvblue]{hyperref}
\usepackage{soul}
\usepackage{multirow}
\usepackage{xcolor}

\def\wacvPaperID{1216} % *** Enter the WACV Paper ID here
\def\confName{WACV}
\def\confYear{2027}

\title{HiVE-VQA: Hierarchical Interaction via Volumetric Embedding for Video Quality Assessment}

\author{
Robbie Hamilton, Fan Zhang, and David Bull\\
Visual Information Laboratory, University of Bristol, Bristol, UK\\
{\tt\small \{robbie.hamilton, fan.zhang, dave.bull\}@bristol.ac.uk}
}

\begin{document}
\maketitle
\begin{abstract}
The proliferation of user-generated video content has made accurate and robust no-reference video quality assessment (NR-VQA) a highly important challenge. Existing research increasingly relies on architecturally complex multi-branch pipelines or computationally intensive fine-tuning of single-branch backbones. In this paper, we investigate whether a single frozen video foundation model can serve as an effective NR-VQA backbone, and propose HiVE-VQA, the first framework (to our knowledge) to apply a JEPA-based video encoder to this task. HiVE-VQA extracts token grids from multiple depths of a pre-trained V-JEPA 2.1 encoder and stacks them into a hierarchical spatio-temporal volume, which a compact 3D convolutional head fuses across its layer, spatial, and temporal axes, combining low-level distortion cues from shallow layers with the semantic and temporal context captured at greater depths. The encoder remains frozen throughout, so only the head is trained: 5.2M parameters (5.7\% of the total), with no auxiliary multi-modal models and an inference cost that is invariant to video resolution. Trained solely on LSVQ and evaluated without any fine-tuning, HiVE-VQA outperforms 17 state-of-the-art (SOTA) methods in SROCC and PLCC on both intra-dataset (LSVQ-test, LSVQ-1080p) and cross-dataset (KoNViD-1k, LIVE-VQC) benchmarks. These results indicate that hierarchical, depth-wise fusion of features from a single frozen video foundation model provides a more effective and efficient basis for NR-VQA than task-specific multi-network fusion. Code and model checkpoints are available at \url{https://github.com/RobHam99/HiVE-VQA}.
\end{abstract}
\begin{figure}
    \centering
    \includegraphics[width=1.0\linewidth]{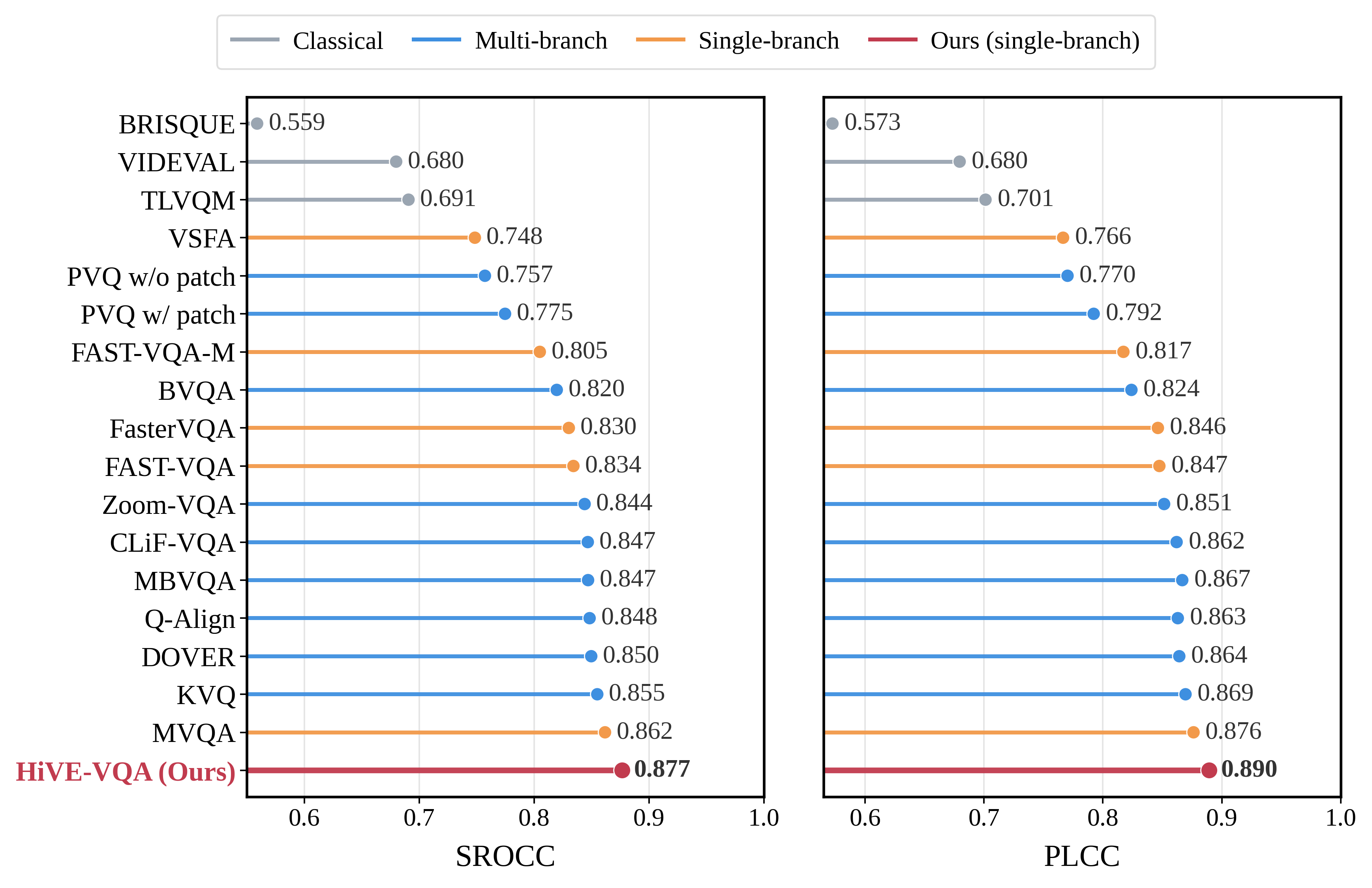}
    \caption{The performance of the proposed HiVE-VQA model, and selected benchmark NR-VQA methods. Here, SROCC and PLCC are used to measure each model's correlation with human-based mean opinion scores, and the results are averaged across the four test datasets. HiVE-VQA is the top performing model compared to 17 classical, deep (single-branch) and deep (multi-branch) models. The results shown here are based on frozen model parameters, without performing any fine-tuning on these datasets.}
    \label{fig:results}
\end{figure}
\section{Introduction}
With the emergence of social networking sites in the previous decades, there has been a significant increase in the amount of video content produced and consumed across the world every single day. According to the CISCO annual report \cite{cisco2020air}, $82\%$ of global internet traffic is video content, and the number of mobile devices grew from 8 billion in 2018, to 13 billion in 2023. This rapid adoption of portable handheld devices, combined with the increase in user-generated content (UGC) produced on social networking sites, creates a huge demand for automated systems that are capable of monitoring and evaluating the quality of content produced and uploaded to these sites. While humans are excellent at evaluating video quality, these subjective evaluations are time consuming and costly. It is of vital importance to develop objective video quality assessment (VQA) metrics that are capable of rapidly and accurately assessing the quality of this user-generated content.

Video quality assessment metrics are typically divided into two sub-categories: conventional and deep-learning metrics. Conventional VQA metrics \cite{brisque,rapique,tlvqm, videval, bliinds, viideo, chipqa, niqe, vmaf} exploit the statistical properties of video scenes to handcraft features that correlate with perceptual quality. These features can be modeled directly, or using machine learning-based models such as random forests. Due to the rigidity of these techniques, they often fail to capture the information required to model complex combinations of distortions and content types. Consequently, these models may generalize poorly to real-world UGC, which exhibits a diverse range of content and distortion characteristics.

The advance of deep learning architectures over the past decade has stimulated substantial interest in their application to VQA. Deep VQA metrics \cite{patchvq, bvqa, coinvq, dover, maxvqa, clifvqa, mbvqa, vsfa, gst-vqa, fast-vqa, faster-vqa,mvqa, rankdvqa, rankdvqamini, conviqt, contrique, qalign, zoomvqa} have demonstrated a strong capability in extracting high-dimensional features that capture diverse distortions with different content types. These models have achieved excellent performance in full-reference (FR) tasks - where the features of a distorted video can be directly compared with those of its pristine reference. However, UGC typically lacks such pristine counterparts, since their degradations may originate during capture due to limitations in recording devices, acquisition conditions, or film practices \cite{li2021user}. Consequently, no-reference (NR) quality assessment for UGC poses a more challenging problem, requiring models to infer perceptual quality solely from the degraded video itself.

Of relevance to this challenge, a new self-supervised deep-learning framework, Joint-Embedding Predictive Architecture (JEPA) \cite{jepa}, has been designed to learn semantically rich and context-aware image representations for downstream tasks. Recent works have extended the JEPA paradigm to video \cite{vjepa2} and further enhanced local spatio-temporal representations through dense predictive objectives \cite{vjepa2_1}, making JEPA-based models promising backbone candidates for low-level vision tasks.

In this context, we investigate whether a frozen, self-supervised video foundation model can serve as an effective backbone for NR-VQA without task-specific encoder fine-tuning or multi-network ensembles. The paper proposes HiVE-VQA, which first extracts tokens from multiple depths of a frozen V-JEPA 2.1 encoder in order to capture dense multi-scale features. HiVE-VQA then restructures the tokens into a hierarchical volume - combining depth, spatial, and temporal dimensions. This restructuring process allows us to apply lightweight 3D convolutions across the volume, mapping our spatio-temporal and multi-scale features into a single quality score. The main contributions of our work are:

\begin{itemize}
    \item \textbf{JEPA backbone for NR-VQA.} To the best of our knowledge, this is the first application of a JEPA-based video encoder to NR-VQA. We show that a frozen V-JEPA 2.1 backbone provides a substantially stronger fixed representation than a comparably sized image-pretrained ViT (+0.080 average SROCC, Table~\ref{tab:regressor_ablation}), requiring no encoder fine-tuning and only 5.2M trainable parameters.
    \item \textbf{A volumetric quality head.} We propose a compact 3D convolutional head that takes restructured token grids at multiple depths of the vision transformer and then fuses them across the depth, spatial, and temporal axes. It outperforms a baseline regressor by +0.040 average SROCC (Table~\ref{tab:regressor_ablation}), while hierarchical multi-layer fusion provides a further improvement over final-layer features at negligible additional cost (Table~\ref{tab:layer_ablation}). 
    \item \textbf{State-of-the-art accuracy.} HiVE-VQA achieves the best performance across four UGC benchmarks (Figure~\ref{fig:results}) among 17 existing SOTA NR-VQA methods, supporting resolution-invariant inference.
\end{itemize}
\begin{figure*}[t]
    \centering
    \includegraphics[width=1.0\linewidth]{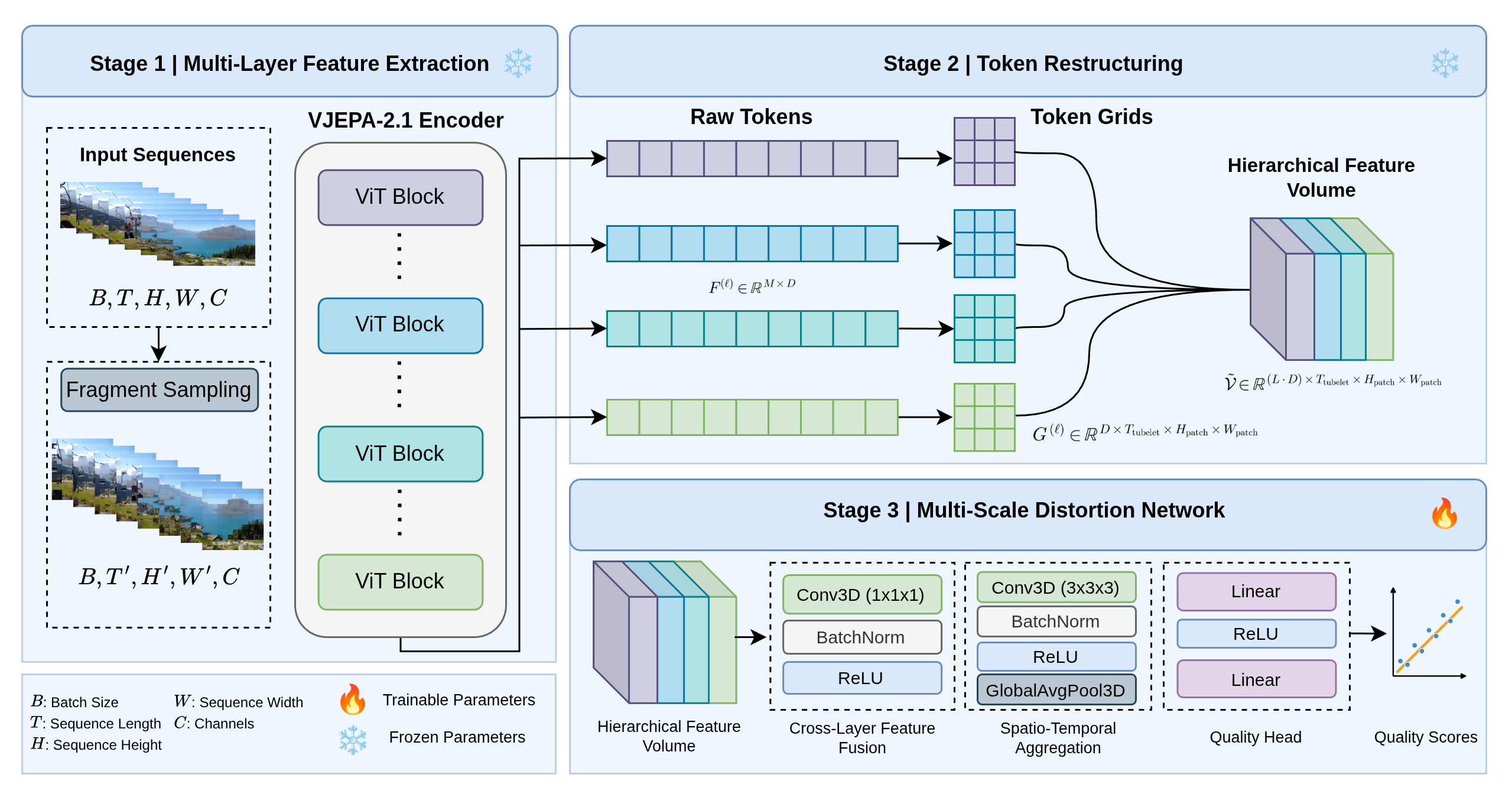}
    \caption{The framework of the proposed HiVE-VQA method. 
    %First, the input video sequences are sampled and fed into the backbone, where tokens from uniformly sampled layers are extracted. The resulting token sequences are subsequently restructured into a hierarchical volumetric embedding. Finally, 3D convolutions are applied to this embedding to extract multi-scale quality features, which  are then mapped to perceptual quality scores.
    }
    \label{fig:architecture}
\end{figure*}

\section{Related Work}
\subsection{Video Quality Assessment}

The lack of a pristine reference sequence makes no-reference assessment inherently more challenging than full-reference tasks, as models cannot directly exploit the statistical or representational differences between distorted and reference content. No-Reference VQA methods are typically divided into two categories: classical and deep learning-based. 

\vspace{5pt}\noindent\textbf{Classical NR-VQA methods.}
Many conventional image quality metrics are based on measuring the similarity of handcrafted frame-level features to the distributions of natural scene statistics. These frame-level features are then pooled along the temporal dimension and mapped to a quality score using a regression technique, most commonly the support vector regressor \cite{brisque, niqe,vmaf}. To exploit the additional temporal information present in videos, video-specific metrics were subsequently developed. Models like VIIDEO \cite{viideo} leverage frame differentials, while BLIINDS \cite{bliinds} utilizes natural scene statistics with additional temporal modeling. ChipQA \cite{chipqa} is motivated by the concept of space-time slices, measuring naturalness at the video level. Some other conventional metrics reduce complexity by selecting key frames of a video to compute multi-level features on \cite{tlvqm, videval}. While these models employ sophisticated modeling techniques, the diversity of UGC content requires models capable of extracting more dynamic features.

\vspace{5pt}\noindent\textbf{Deep learning-based NR-VQA methods.}
The massive increase in visual data over the past decade, coupled with advances in deep learning, has stimulated the development of deep VQA models. While such metrics are computationally demanding, they typically produce features that are more expressive and sensitive to a wide range of distortions and content types. CNN-based VQA models were among the earliest deep architectures investigated for NR-VQA \cite{rapique, patchvq, gst-vqa, bvqa}. Some have been trained using semi-supervised training paradigms \cite{contrique, conviqt}, allowing the models to learn from a greater volume of data without requiring human-annotated quality labels.
The advent of the transformer architecture \cite{vaswani2017attention, dosovitskiy2020image} has yielded NR-VQA methods with increasingly strong performance. Semi-supervised training schemes have also been used with transformer models \cite{rankdvqa, rankdvqamini}. To counter the computational demands of transformer models, FAST-VQA \cite{fast-vqa, faster-vqa} introduced a fragment sampling strategy, allowing computational costs to remain approximately invariant to the source video resolution.
Many modern NR-VQA metrics trade efficiency for improved performance by integrating multiple specialist networks that capture complementary views, scales, or aspects of video quality \cite{coinvq, dover, mbvqa, zoomvqa, clifvqa}. While transformers and CNNs are still the dominant architectures in modern NR-VQA, other techniques such as language-based models \cite{qalign, maxvqa}, and state-space models \cite{mvqa} are also demonstrating excellent performance on UGC quality assessment tasks.

\section{Proposed Method}

\subsection{Overall Framework}
Figure~\ref{fig:architecture} presents an overview of the proposed HiVE-VQA framework. It
repurposes a self-supervised video encoder, pre-trained with the V-JEPA 2.1
objective \cite{vjepa2_1}, as a hierarchical feature extractor for no-reference
video quality assessment. The framework consists of three stages.

\textbf{(i) Data sampling.} An input sequence
$X\in\mathbb{R}^{T\times H\times W\times C}$ is first mapped to a fixed-size
fragment collage $X'\in\mathbb{R}^{T'\times H'\times W'\times 3}$ using the
fragment sampling strategy introduced in \cite{fast-vqa}, which preserves local textures and
authentic distortions while reducing spatial redundancy. 

\textbf{(ii) Hierarchical feature extraction.} The sampled clip $X'$ is passed
through the frozen V-JEPA 2.1 ViT-B encoder. Rather than using only the final
representation, we extract features from a uniformly spaced subset of $L$ transformer blocks and fold their dense token sequences back into spatio-temporal grids, which are stacked into a single hierarchical feature volume
$\tilde{\mathcal{V}}\in\mathbb{R}^{(L\cdot D)\times T_{\text{tubelet}}\times H_{\text{patch}}\times W_{\text{patch}}}$ (Section~\ref{sec:hvnet}). Shallow blocks capture low-level spatial distortions (blur, blocking, noise), while deeper blocks encode high-level semantic and temporal context, so the volume jointly represents quality cues at multiple depths of the network.

\textbf{(iii) Volumetric quality prediction.} A lightweight 3D convolutional head fuses the hierarchical volume across its layer, spatial, and temporal axes and predicts a single scalar quality score $\hat{s}$ (Section~\ref{sec:head}).

\subsection{Data Sampling}

Given an input sequence $X\in\mathbb{R}^{T\times H \times W\times C}$, the aim of a VQA model is to map this sequence to a scalar quality score. In the proposed HiVE-VQA framework, we first adopt the fragment-sampling strategy introduced in \cite{fast-vqa}, transforming the input sequence, $X$, into a sampled fragment sequence, $X'\in\mathbb{R}^{T'\times H'\times W'\times3}$, where $T'$ is the number of frames sampled from the input video, and $H'=W'$ denotes the spatial dimensions of the resulting fragment collage. Each sampled clip is comprised of fragments of size $f_s\times f_s$, where $f_s$ is a factor of $H'$, i.e., the fragments form a regular square grid. We use $T'=32$ frames and a fragment collage of $H'=W'=384$, assembled from a $12\times 12$ grid of fragments of size $f_s\times f_s=32\times 32$. HiVE-VQA-M uses $T'=16$, $H'=W'=224$ with a $7\times 7$ grid.

\subsection{Hierarchical Feature Extraction}
\label{sec:hvnet}

The sampled input, $X'$, is passed to the first stage of our model, the encoder, which  contains ViT blocks numbered $1, \dots, \text{B}$. We extract a uniformly spaced subset of layers across the network depth, $\mathcal{S}=\{\ell_1, \dots,\ell_L\}$, where $L=4$ is the number of sampled layers. For a sampled clip of input shape $T'\times H'\times W'\times 3$, we extract a dense token sequence from each ViT block:
\begin{equation*}
    F^{(\ell)}\in \mathbb{R}^{M\times D} , \quad  M=T_{\text{tubelet}}\cdot H_{\text{patch}}\cdot W_{\text{patch}},
\end{equation*}
in which $D$ is the ViT embedding dimension and $(T_{tubelet}, H_{patch}, W_{patch})$ denotes the temporal tubelet and spatial-patch grid sizes. Each ViT block in the network shares an identical token and embedding structure, meaning that each layer shares the same spatio-temporal layout and can be reshaped into a spatio-temporal token grid:
\begin{equation*}
    G^{(\ell)}\in\mathbb{R}^{D\times T_{\text{tubelet}}\times H_{\text{patch}}\times W_{\text{patch}}}.
\end{equation*}
By stacking the spatio-temporal token grid along a new dimension, we create a hierarchical feature volume:
\begin{equation*}
    \mathcal{V}\in\mathbb{R}^{L\times D\times T_{\text{tubelet}}\times H_{\text{patch}}\times W_{\text{patch}}}.
\end{equation*}
We then collapse the layer and embedding axes into a single channel dimension:
\begin{equation*}
    \tilde{\mathcal{V}}\in\mathbb{R}^{(L\cdot D)\times T_{\text{tubelet}}\times H_{\text{patch}}\times W_{\text{patch}}}.
\end{equation*}
This hierarchical volume enables the quality prediction head  to simultaneously learn quality mappings from multi-scale spatio-temporal features.

\subsection{Volumetric Quality Prediction}
\label{sec:head}

User-generated content exhibits a wide range of spatial and temporal
distortions that may appear locally, in isolated regions, or globally across a
sequence. To predict quality across multiple scales, we introduce a network that models quality-relevant distortions within the hierarchical feature volume
$\tilde{\mathcal{V}}\in\mathbb{R}^{(L\cdot D)\times T_{\text{tubelet}}\times H_{\text{patch}}\times W_{\text{patch}}}$
using a fully convolutional 3D head, preserving the spatio-temporal layout of
the tokens until the final pooling stage.

\vspace{5pt}\noindent\textbf{Cross-layer fusion.} We first fuse information across the $L$ sampled
layers and $D$ embedding channels at every spatio-temporal location using a
pointwise ($1\times1\times1$) 3D convolution, followed by batch normalization and a ReLU non-linearity:
\begin{equation*}
    Z = \sigma\!\big(\mathrm{BN}(W_{p}\ast\tilde{\mathcal{V}})\big),
    \quad Z\in\mathbb{R}^{C_{1}\times T_{\text{tubelet}}\times H_{\text{patch}}\times W_{\text{patch}}},
\end{equation*}
where $W_{p}$ projects the $(L\cdot D)$ input channels to $C_{1}=512$. This operation
learns a location-specific fusion of the hierarchical features, allowing the
head to emphasize the layers most indicative of quality for a given region.

\vspace{5pt}\noindent\textbf{Spatio-temporal aggregation.} A subsequent $3\times3\times3$
convolution (with $C_{1}\!\to\!C_{2}=256$ channels), batch normalization, and
ReLU aggregate local spatio-temporal neighborhoods, capturing distortions
that span adjacent tubelets and patches. The resulting activations are reduced
to a single descriptor by global average pooling over the temporal and spatial
dimensions:
\begin{equation*}
    z = \mathrm{AvgPool}_{3\mathrm{D}}\!\big(\sigma(\mathrm{BN}(W_{c}\ast Z))\big)
    \in\mathbb{R}^{C_{2}}.
\end{equation*}
Aggregating quality evidence only at the final stage allows the network to form a
holistic judgment that integrates both localized and global degradations,
rather than predicting region-level quality scores at intermediate stages.

\vspace{5pt}\noindent\textbf{Regression.} Finally, a two-layer multilayer perceptron with a ReLU
activation maps the pooled descriptor to a scalar quality score:
\begin{equation*}
    \hat{s} = W_{2}\,\sigma(W_{1}z),
    \qquad W_{1}\in\mathbb{R}^{128\times C_{2}},\; W_{2}\in\mathbb{R}^{1\times128}.
\end{equation*}
The entire head contains only a few million parameters, making it more efficient to train compared to large scale end-to-end fine-tuning paradigms.

\subsection{The Hybrid Ranking-correlation Loss Function}
\label{sec:loss}

Our network is trained with a hybrid loss function configured to optimize both the ranking monotonicity and linear correlation with human perceptual quality. Our loss is inspired by other hybrid loss functions used in the literature \cite{mvqa, zoomvqa}.
\begin{equation}
\mathcal{L}_{\text{hybrid}}=\omega_{\text{rank}}\mathcal{L}_{\text{rank}}+\omega_{\text{PLCC}}\mathcal{L}_{\text{PLCC}},
\label{eq:hybrid_loss}
\end{equation}
where the ranking-based loss function $\mathcal{L}_{\text{rank}}$ is designed to guide the model to learn the quality monotonicity of user-generated content. For all valid pairs $(i,j)$ such that $y_i\neq y_j$, we define:
\begin{equation}
    s_{ij}=\text{sign}(y_i-y_j),
\end{equation}
and prediction differences:
\begin{equation}
    d_{ij}=\hat{y}_i-\hat{y}_j.
\end{equation}
The margin ranking loss is:
\begin{equation}
    \mathcal{L}_{\text{rank}}=\frac{1}{|\mathcal{P}|}\sum_{(i,j)\in\mathcal{P}}\text{max}(0,-s_{ij}d_{ij}+m),
\end{equation}
where $m$ is the ranking margin, while $\mathcal{P}$ denotes the set of unequal score pairs.

Differentiable Pearson Linear Correlation Coefficient (PLCC) losses are commonplace in the quality assessment field \cite{li2020norm, li_unified_2021, ma2018geometric, liu2018end}. This component teaches the model to linearly align its predictions with human perceptual ratings. Here, we normalize both predictions and targets to zero mean and unit variance:
\begin{equation}
\tilde{y}_i
=
\frac{\hat y_i - \mu_{\hat y}}
{\sigma_{\hat y} + \varepsilon}, \quad
\tilde{t}_i
=
\frac{y_i - \mu_y}
{\sigma_y + \varepsilon},
\end{equation}
where
\begin{equation}
\mu_{\hat y} = \frac{1}{N}\sum_{i=1}^N \hat y_i,
\qquad
\sigma_{\hat y}
=
\sqrt{
\frac{1}{N}
\sum_{i=1}^N
(\hat y_i-\mu_{\hat y})^2
},
\end{equation}
and similarly for \( \mu_y \) and \( \sigma_y \).
Here, \( \varepsilon \) is a small constant for numerical stability.

The PLCC between normalized predictions and targets is computed as:
\begin{equation}
\rho
=
\frac{1}{N}
\sum_{i=1}^N
\tilde y_i \tilde t_i.
\end{equation}

The PLCC-based loss contains two terms.
The first term minimizes the normalized mean squared error:
\begin{equation}
\mathcal{L}_{\mathrm{MSE}}
=
\frac{1}{4N}
\sum_{i=1}^N
(\tilde y_i - \tilde t_i)^2.
\end{equation}
The second term aligns the prediction scale with the correlation coefficient:
\begin{equation}
\mathcal{L}_{\mathrm{corr}}
=
\frac{1}{4N}
\sum_{i=1}^N
(\rho \tilde y_i - \tilde t_i)^2.
\end{equation}

The final PLCC loss is defined as:
\begin{equation}
\mathcal{L}_{\mathrm{PLCC}}
=
\frac{
\mathcal{L}_{\mathrm{MSE}}
+
\mathcal{L}_{\mathrm{corr}}
}{2}.
\end{equation}

\section{Experiments}

\begin{table*}[t]
\centering
\caption{Evaluation of the proposed HiVE-VQA methods and state-of-the-art NR-VQA metrics on UGC video quality datasets. \textcolor{red}{Red} corresponds to best, \textcolor{blue}{blue} corresponds to second best and \textcolor{teal}{teal} corresponds to third best (excluding HiVE-VQA-M).}
\label{table:pretraining}
\resizebox{\textwidth}{!}{
\begin{tabular}{rrrccccccccccc}
\toprule
\multicolumn{3}{c}{Testing Type} & \multicolumn{4}{c}{Intra-Dataset} & \multicolumn{4}{c}{Cross-Dataset} & \multicolumn{2}{c}{Average} \\
\midrule
\multicolumn{3}{c}{Testing Datasets} 
& \multicolumn{2}{c}{LSVQ\textsubscript{test}} 
& \multicolumn{2}{c}{LSVQ\textsubscript{1080p}}
& \multicolumn{2}{c}{KoNViD-1k}
& \multicolumn{2}{c}{LIVE-VQC}
& \multicolumn{2}{c}{Overall} \\
\midrule
Type & Methods & Source
& SROCC & PLCC 
& SROCC & PLCC 
& SROCC & PLCC 
& SROCC & PLCC 
& SROCC & PLCC \\
\midrule
\multirow{3}{*}{Classical}
& BRISQUE \cite{brisque} & TIP, 2012 & 0.569 & 0.576 & 0.497 & 0.531 & 0.646 & 0.647 & 0.524 & 0.536 & 0.559 & 0.573 \\
& TLVQM \cite{tlvqm} & TIP, 2019 & 0.772 & 0.774 & 0.589 & 0.616 & 0.732 & 0.724 & 0.670 & 0.691 & 0.691 & 0.701 \\
& VIDEVAL \cite{videval} & TIP, 2021 & 0.794 & 0.783 & 0.545 & 0.554 & 0.751 & 0.741 & 0.630 & 0.640 & 0.680 & 0.680 \\
\midrule
\multirow{8}{*}{Deep (Multi-branch)}
& PVQ w/o patch \cite{patchvq}& CVPR, 2021 & 0.814 & 0.816 & 0.686 & 0.708 & 0.782 & 0.781 & 0.747 & 0.776 & 0.757 & 0.770 \\
& PVQ w/ patch \cite{patchvq} & CVPR, 2021 & 0.827 & 0.828 & 0.711 & 0.739 & 0.791 & 0.795 & 0.770 & 0.807 & 0.775 & 0.792 \\
& BVQA \cite{bvqa}& TCSVT, 2022 & 0.852 & 0.854 & 0.772 & 0.788 & 0.839 & 0.830 & 0.816 & 0.824 & 0.820 & 0.824 \\
& DOVER \cite{dover}& ICCV, 2023 & 0.888 & 0.889 & 0.795 & 0.830 & 0.884 & 0.883 & 0.832 & \textcolor{teal}{0.855} & 0.850 & 0.864 \\
& Zoom-VQA \cite{zoomvqa} & CVPRW, 2023 & 0.886 & 0.879 & 0.799 & 0.819 & 0.877 & 0.875 & 0.814 & 0.833 & 0.844 & 0.852 \\
& Q-Align \cite{qalign} & ICML, 2024 & 0.883 & 0.882 & 0.797 & 0.830 & 0.865 & 0.877 & -- & -- & 0.848 & 0.863 \\
& CLiF-VQA \cite{clifvqa} & ACMMM, 2024 & 0.886 & 0.887 & 0.790 & 0.832 & 0.877 & 0.874 & \textcolor{teal}{0.834} & \textcolor{teal}{0.855} & 0.847 & 0.862 \\
& MBVQA \cite{mbvqa} & CVPR, 2024 & 0.895 & 0.895 & 0.809 & 0.844 & 0.878 & 0.884 & 0.806 & 0.844 & 0.847 & 0.867 \\
& KVQ \cite{qu2025kvq} & CVPR, 2025
& \textcolor{teal}{0.896} & \textcolor{teal}{0.897}
& \textcolor{blue}{0.814} & \textcolor{blue}{0.846} 
& \textcolor{blue}{0.890} & \textcolor{blue}{0.892} 
& 0.820 & 0.843
& \textcolor{teal}{0.855} & \textcolor{teal}{0.870} \\
\midrule
\multirow{6}{*}{Deep (Single-branch)}
& VSFA \cite{vsfa} & ACMMM, 2019 & 0.801 & 0.796 & 0.675 & 0.704 & 0.784 & 0.794 & 0.734 & 0.772 & 0.749 & 0.767 \\
& FAST-VQA-M \cite{fast-vqa} & ECCV, 2022 & 0.852 & 0.854 & 0.739 & 0.773 & 0.841 & 0.832 & 0.788 & 0.810 & 0.805 & 0.817 \\
& FAST-VQA \cite{fast-vqa} & ECCV, 2022 & 0.876 & 0.877 & 0.779 & 0.814 & 0.859 & 0.855 & 0.823 & 0.844 & 0.834 & 0.848 \\
& FasterVQA \cite{faster-vqa} & TPAMI, 2023 & 0.873 & 0.874 & 0.772 & 0.811 & 0.863 & 0.863 & 0.813 & 0.837 & 0.830 & 0.846 \\
& MVQA \cite{mvqa} & ICCV, 2025 
& \textcolor{blue}{0.898} & \textcolor{blue}{0.899} 
& \textcolor{teal}{0.812} & \textcolor{blue}{0.846} 
& \textcolor{teal}{0.885} & \textcolor{teal}{0.887}
& \textcolor{blue}{0.852} & \textcolor{blue}{0.873} 
& \textcolor{blue}{0.862} & \textcolor{blue}{0.876} \\
\midrule
& HiVE-VQA-M (ours) & -- &
0.891 & 0.891 &
0.819 & 0.850 & 
0.880 & 0.881 &
0.835 & 0.860 & 
0.856 & 0.871 \\
& HiVE-VQA (ours) & -- 
& \textcolor{red}{0.906} & \textcolor{red}{0.906} &
\textcolor{red}{0.835} & \textcolor{red}{0.867} & 
\textcolor{red}{0.902} & \textcolor{red}{0.903} & 
\textcolor{red}{0.864} & \textcolor{red}{0.882} &
\textcolor{red}{0.877} & \textcolor{red}{0.890} \\
\midrule
\multicolumn{3}{l}{Improvement over best Multi-branch} 
& +1.11\% & +1.00\% & +2.58\% & +2.48\% & +1.35\% & +1.23\% & +3.60\% & +3.16\% & +2.57\% & +2.30\% \\
\midrule
\multicolumn{3}{l}{Improvement over best Single-branch} 
& +0.89\% & +0.78\% & +2.83\% & +2.48\% & +1.92\% & +1.80\% & +1.41\% & +1.03\% & +1.74\% & +1.60\% \\
\bottomrule
\end{tabular}
}
\end{table*}
\begin{figure*}[t]
    \centering
    \includegraphics[width=1.0\linewidth]{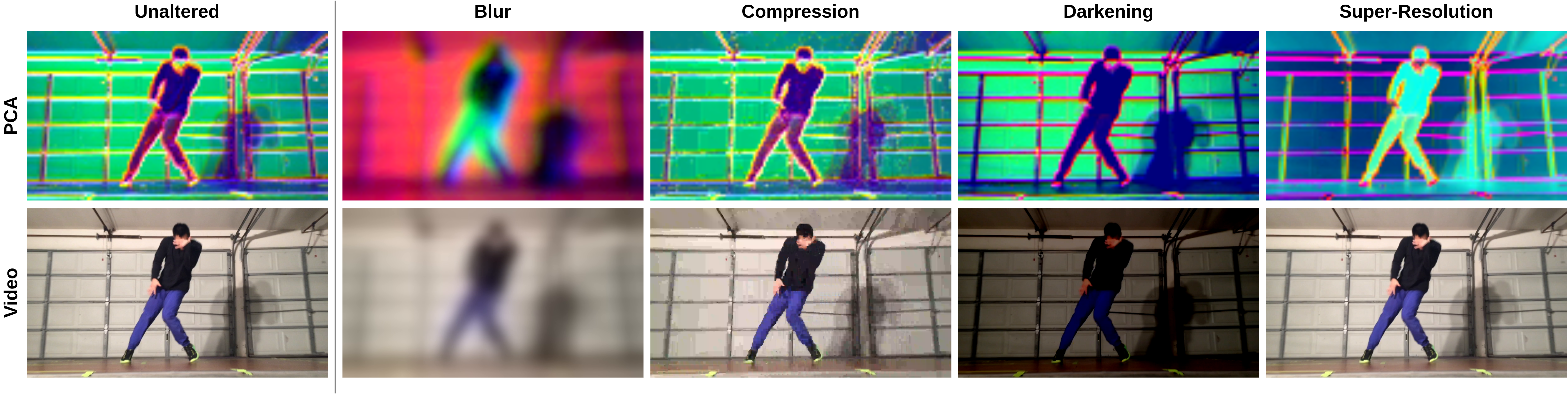}
    \caption{\textbf{Representation of artifacts}. We apply 4 synthetic distortions (blur, compression, darkening, super-resolution) to a source sequence, in order to determine if the encoder is able to \textit{see} artifacts. We perform principal component analysis (PCA) on output features from the V-JEPA 2.1 encoder; the first three principal components are then mapped to the RGB channels.} 
    \label{fig:pca}
\end{figure*}

%\begin{table}
%\caption{UGC VQA Datasets}
%    \centering
%    \begin{tabular}{cccc}
%    \hline
%         Dataset & Task & Scenes & Resolution \\
%         \hline
%         KoNViD-1k \cite{konvid1k, hosu2017konstanz} & VQA & 1,200 & 540p \\
%         LIVE-VQC \cite{livevqc} & VQA & 585 & 240p-1080p \\
%         LSVQ \cite{patchvq} & VQA & 39,076 & 360p-1080p\\
%         YouTube-UGC \cite{youtubeugc} & VQA & 1,380 & 360p-1080p\\
%          \hline
%    \end{tabular}
%    
%    \label{tab:datasets}
%\end{table}

\begin{table*}[t]
\centering
\caption{Fine-tuning results across multiple UGC datasets. \textcolor{red}{Red} corresponds to best, \textcolor{blue}{blue} corresponds to second best  and \textcolor{teal}{teal} corresponds to third best (excluding HiVE-VQA-M). DVLTA-VQA* \cite{dvlta-vqa} was not included in Table~\ref{table:pretraining}, as their code has not yet been released.}
\resizebox{\textwidth}{!}{
\begin{tabular}{r r l cc cc cc cc}
\toprule
\multirow{2}{*}{Type} 
& \multirow{2}{*}{Methods} 
& \multirow{2}{*}{Source}
& \multicolumn{2}{c}{LIVE-VQC}
& \multicolumn{2}{c}{KoNViD-1k}
& \multicolumn{2}{c}{YouTube-UGC}
& \multicolumn{2}{c}{Average} \\
\cmidrule(lr){4-11}
& & 
& SROCC & PLCC
& SROCC & PLCC
& SROCC & PLCC
& SROCC & PLCC \\
\midrule
\multirow{3}{*}{Classical}
& TLVQM~\cite{tlvqm} & TIP, 2019
& 0.799 & 0.803
& 0.773 & 0.768
& 0.669 & 0.659
& 0.732 & 0.726 \\
& VIDEVAL~\cite{videval} & TIP, 2021
& 0.752 & 0.751
& 0.783 & 0.780
& 0.779 & 0.773
& 0.772 & 0.772 \\
& RAPIQUE~\cite{rapique} & OJSP, 2021
& 0.755 & 0.786
& 0.803 & 0.817
& 0.759 & 0.768
& 0.774 & 0.790 \\
\midrule
\multirow{2}{*}{Classical + Deep}
& CNN+TLVQM~\cite{cnntlvqm} & ACMMM, 2020
& 0.825 & 0.834
& 0.816 & 0.818
& N/A & N/A
& 0.821 & 0.826 \\
& CNN+VIDEVAL~\cite{videval} & TIP, 2021
& 0.785 & 0.810
& 0.815 & 0.817
& 0.808 & 0.803
& 0.806 & 0.810 \\
\midrule
\multirow{7}{*}{Deep (Multi-branch)}
& PVQ~\cite{patchvq} & CVPR, 2021
& 0.827 & 0.837
& 0.791 & 0.786
& N/A & N/A
& N/A & N/A \\
& BVQA~\cite{bvqa} & TCSVT, 2022
& 0.831 & 0.842
& 0.834 & 0.836
& 0.831 & 0.819
& 0.832 & 0.832 \\
& CoINVQ~\cite{coinvq} & CVPR, 2021
& N/A & N/A
& 0.767 & 0.764
& 0.816 & 0.802
& N/A & N/A \\
& DOVER~\cite{dover} & ICCV, 2023
& 0.860 & 0.875
& 0.909 & 0.906
& 0.890 & 0.891
& 0.886 & 0.891 \\
& MaxVQA~\cite{maxvqa} & ACMMM, 2023
& 0.854 & 0.873
& 0.894 & 0.895
& 0.894 & 0.890
& 0.881 & 0.886 \\
& CLiF-VQA~\cite{clifvqa} & ACMMM, 2024
& 0.866 & 0.878
& 0.903 & 0.903
& 0.888 & 0.890
& 0.886 & 0.890 \\
& MBVQA~\cite{mbvqa} & CVPR, 2024
& 0.860 & 0.880
& 0.901 & 0.905
& 0.876 & 0.877
& 0.879 & 0.887 \\
& KVQ~\cite{qu2025kvq} & CVPR, 2025
& 0.859 & 0.879
& 0.909 & \textcolor{teal}{0.915}
& \textcolor{blue}{0.903} & \textcolor{red}{0.905}
& 0.890 & 0.900 \\
& DVLTA-VQA* \cite{dvlta-vqa} & TCSVT, 2026
& \textcolor{blue}{0.892} & \textcolor{blue}{0.903}
& \textcolor{teal}{0.912} & 0.913
& \textcolor{red}{0.918} & \textcolor{red}{0.905}
& \textcolor{blue}{0.907} & \textcolor{teal}{0.907} \\
\midrule
\multirow{7}{*}{Deep (Single-branch)}
& VSFA~\cite{vsfa} & ACMMM, 2019
& 0.773 & 0.795
& 0.773 & 0.775
& 0.724 & 0.743
& 0.752 & 0.765 \\
& FAST-VQA-M~\cite{fast-vqa} & ECCV, 2022
& 0.803 & 0.828
& 0.873 & 0.872
& 0.768 & 0.765
& 0.815 & 0.822 \\
& FAST-VQA~\cite{fast-vqa} & ECCV, 2022
& 0.845 & 0.852
& 0.890 & 0.889
& 0.857 & 0.853
& 0.864 & 0.865 \\
& FasterVQA~\cite{faster-vqa} & TPAMI, 2023
& 0.843 & 0.858
& 0.895 & 0.898
& 0.863 & 0.859
& 0.867 & 0.872 \\
& MVQA \cite{mvqa} & ICCV, 2025
& \textcolor{teal}{0.878} & \textcolor{teal}{0.895}
& \textcolor{red}{0.925} & \textcolor{red}{0.925}
& \textcolor{teal}{0.901} & \textcolor{teal}{0.903}
& \textcolor{teal}{0.901} & \textcolor{blue}{0.908} \\
\midrule
& HiVE-VQA-M & Ours
& 0.877 & 0.892
& 0.905 & 0.904
& 0.823 & 0.829
& 0.868 & 0.875 \\
& HiVE-VQA & Ours
& \textcolor{red}{0.910} & \textcolor{red}{0.926}
& \textcolor{blue}{0.921} & \textcolor{red}{0.925}
& 0.894 & 0.899
& \textcolor{red}{0.908} & \textcolor{red}{0.917} \\
\midrule
\multicolumn{3}{l}{Improvement over existing best Single-branch}
& +3.64\% & +3.46\% & -0.43\% & +0.00\% & -0.78\% & -0.44\% & +0.78\% & +1.00\% \\
\multicolumn{3}{l}{Improvement over existing best Multi-branch}
& +2.02\% & +2.55\% & +0.99\% & +1.09\% & -2.61\% & -0.66\% & +0.11\% & +1.10\% \\
\bottomrule
\end{tabular}
}
\label{tab:finetuning}
\end{table*}

\begin{table}[t]
\centering
\caption{Details of the HiVE-VQA models.}
\resizebox{\columnwidth}{!}{
\begin{tabular}{l c c c c c}
\toprule
Models & Input Size & Depth & Dim & Param & FLOPs \\
\midrule
HiVE-VQA-M
& $16 \times 224 \times 224$ 
& 12
& 768
& 91M
& 188G \\
HiVE-VQA
& $32 \times 384 \times 384$ 
& 12
& 768
& 91M
& 2406G \\
\bottomrule
\end{tabular}
}
\label{tab:model_params}
\end{table}

\subsection{Experimental Settings}
\vspace{5pt}\noindent\textbf{UGC Datasets.} We evaluate the proposed HiVE-VQA variants on four widely used UGC subjective quality datasets, including $\text{LSVQ}_{\text{1080p}}$ and $\text{LSVQ}_{\text{test}}$ \cite{patchvq}, KoNViD-1k \cite{konvid1k, hosu2017konstanz}, LIVE-VQC \cite{livevqc}, and YouTube-UGC \cite{youtubeugc}. HiVE-VQA is pre-trained on $\text{LSVQ}_{\text{train}}$, which is a subset of the main LSVQ dataset comprising 28,056 video clips. After pre-training, intra-dataset evaluation is performed on the two LSVQ test datasets, which contain 7,400 and 3,600 videos, respectively. To evaluate our model's generalization ability on UGC with distributions different from that of the training data, we perform cross-dataset evaluation on KoNViD-1k (1,200 sequences) and LIVE-VQC (585 sequences). To further assess the model's adaptability to specific data, we also fine-tuned HiVE-VQA on KoNViD-1k, LIVE-VQC, and the public YouTube-UGC dataset (1,380 sequences).

\vspace{5pt}\noindent\textbf{Evaluation Criteria.} The correlation between the predicted quality indices and subjective quality scores is measured based on two correlation statistical metrics: Spearman Rank Order Correlation Coefficient (SROCC) \cite{spearman1904proof} and Pearson Linear Correlation Coefficient (PLCC) \cite{pearson1895note}. The SROCC measures the monotonic relationship between the model's predictions and human-annotated mean-opinion scores (MOS), thereby reflecting the ranking capability of the model. PLCC is used to evaluate the model's prediction accuracy by calculating the linear correlation between the model's predicted scores and the MOS scores after non-linear fitting \cite{vqeg2000}. For both metrics, a higher score indicates better performance.

\vspace{5pt}\noindent\textbf{Training Details.}
We use the PyTorch framework and 8 NVIDIA H200 GPUs to train our model; the model is also compatible with training on a single consumer-grade GPU (NVIDIA RTX 4090). We utilize a pre-trained V-JEPA 2.1 ViT-B model as our vision backbone \cite{vjepa2_1}. We set the initial learning rate to $1e^{-3}$ and use the AdamW optimizer with the ReduceLROnPlateau scheduler. Detailed model parameters are provided in Table~\ref{tab:model_params}. We set the batch size to 32 (per GPU) and weight decay to $1e^{-4}$. As we are training the models with linear and monotonic correlation in mind, we treat each objective with equal importance, and set the hybrid loss function to have equal weighting ($\omega_{\text{rank}}=\omega_{\text{PLCC}}=0.5$) and the default margin $m=1.0$.

\subsection{Pre-training results on LSVQ}
We pre-train our model on LSVQ\textsubscript{train} and conduct inference on the two intra-dataset test sets (LSVQ\textsubscript{test} and LSVQ\textsubscript{1080p}). We conduct additional cross-dataset inference on KoNViD-1k and LIVE-VQC in order to assess the generalizability of our model to UGC. Our results are compared in Table~\ref{table:pretraining} with 17 SOTA classical and deep model benchmarks. Across all four test datasets, our model outperforms all of the classical VQA metrics included in the benchmark. Compared with all single-branch models in the literature, it achieves superior performance, improving mean SROCC and PLCC by 1.74\% and 1.60\%, respectively. Although our model adopts a single-branch architecture, it still outperforms all multi-branch models, with improvements of 2.57\% and 2.30\% in SROCC and PLCC, respectively.

\subsection{Fine-tuning results on UGC datasets}
To verify the effectiveness of our model on downstream tasks, we fine-tune HiVE-VQA on three of the previously mentioned datasets (LIVE-VQC, KoNViD-1k and YouTube-UGC). We repeat the fine-tuning process 10 times across a 70/10/20\% dataset split (train/val/test), and report the median performance across the 10 runs in Table~\ref{tab:finetuning}. It is noted that HiVE-VQA once again achieves the best average performance among all models, improving SROCC and PLCC by 0.11\% and 1.10\%, respectively, over the best multi-branch model. HiVE-VQA also demonstrates competitive fine-tuning performance compared with other state-of-the-art single-branch models. While our model obtains evident performance gains ($+3.64\%$ in SROCC and $+3.46\%$ in PLCC) on LIVE-VQC, it does not consistently outperform all single-branch models when fine-tuned on multiple datasets, and is only the fourth best model on YouTube-UGC \cite{youtubeugc}. This is likely because we keep our encoder frozen, while other networks are able to tune their features to better represent the dataset distribution; the gap widens for HiVE-VQA-M, most visibly on high-resolution YouTube-UGC.

\subsection{Efficiency}

We evaluate the efficiency of HiVE-VQA by comparing both variants with other SOTA NR-VQA models. At 1080p HiVE-VQA is roughly $11\times$, $15\times$, and $29\times$ less computationally expensive than VSFA, PVQ, and BVQA respectively, while processing each video in under a quarter of a second. Compared with more recent fragment-based and efficient models, HiVE-VQA can be considered in the same lightweight regime. Although FasterVQA, FAST-VQA, and DOVER report lower FLOPs, they employ considerably smaller backbones; HiVE-VQA instead retains a full transformer encoder with superior SROCC and PLCC, and only 5.2M trainable parameters. Our smaller variant, HiVE-VQA-M, is on the same order of FLOPs as the most lightweight model, FasterVQA, while yielding average SROCC and PLCC gains of +0.026 and +0.025 respectively. This analysis indicates HiVE-VQA is a promising alternative for realistic implementations of UGC NR-VQA. We acknowledge the larger size of the total parameter count, which future work will aim to condense through techniques such as knowledge distillation \cite{hinton2015distilling} and model pruning \cite{zhu2021vision, lecun1989optimal}.

\begin{table}[t]
\centering
\caption{Parameter counts, FLOPs and running time (average of 10 runs)
on GPU (RTX 4090). Parameters are in M, FLOPs in G, and time in s.
``Trainable'' denotes parameters updated during training; methods that
fine-tune end-to-end have trainable equal to total. Counts were computed
from publicly released checkpoints where not stated in the original paper.}
\label{tab:flops_runtime}
\resizebox{\columnwidth}{!}{%
\begin{tabular}{r cc cc cc cc}
\toprule
\multirow{2}{*}{Methods}
& \multicolumn{2}{c}{Params (M)}
& \multicolumn{2}{c}{540p}
& \multicolumn{2}{c}{720p}
& \multicolumn{2}{c}{1080p} \\
\cmidrule(lr){2-3}\cmidrule(lr){4-5}\cmidrule(lr){6-7}\cmidrule(lr){8-9}
& Trainable & Total
& GFLOPs & Time
& GFLOPs & Time
& GFLOPs & Time \\
\midrule
VSFA~\cite{vsfa}
& \textbf{0.5} & \textbf{24.1}
& 6440 & 1.506 & 11426 & 2.556 & 25712 & 5.291 \\
PVQ~\cite{patchvq}
& 54.2 & 54.2
& 9203 & 1.792 & 13842 & 2.968 & 36760 & 6.556 \\
BVQA~\cite{bvqa}
& 57.8 & 57.8
& 17705 & 3.145 & 31533 & 7.813 & 70714 & 14.34 \\
FAST-VQA~\cite{fast-vqa}
& 27.7 & 27.7
& 284 & 0.056 & 284 & 0.056 & 284 & 0.056 \\
FasterVQA~\cite{faster-vqa}
& 27.7 & 27.7
& \textbf{70} & 0.032 & \textbf{70} & 0.032 & \textbf{70} & 0.032 \\
DOVER~\cite{dover}
& 56.2 & 56.2
& 282 & 0.061 & 282 & 0.061 & 282 & 0.062 \\
CLiF-VQA~\cite{clifvqa}
& N/A & N/A
& 1432 & 0.522 & 1432 & 0.522 & 1432 & 0.522\\
MBVQA~\cite{mbvqa}
& 87.8 & 87.8
& 912 & 0.194 & 1232 & 0.267 & 2150 & 0.891 \\
MVQA~\cite{mvqa}
& 74.0 & 74.0
& 403 & 0.130 & 403 & 0.130 & 403 & 0.130 \\
\midrule
\textbf{HiVE-VQA-M}
& \textbf{5.2} & \textbf{91.4}
& \textbf{188} & \textbf{0.025} & \textbf{188} & \textbf{0.025} & \textbf{188} & \textbf{0.025} \\
\textbf{HiVE-VQA}
& \textbf{5.2} & \textbf{91.4}
& \textbf{2406} & \textbf{0.206} & \textbf{2406} & \textbf{0.206} & \textbf{2406} & \textbf{0.206} \\
\bottomrule
\end{tabular}%
}
\end{table}

\begin{table*}[t]
\centering
\caption{Layer selection analysis. Checkmarks indicate which encoder
blocks were included; the final row is the proposed configuration.
Best result per column in bold.}
\label{tab:layer_ablation}
\resizebox{1\linewidth}{!}{%
\begin{tabular}{cccc c cc cc cc cc cc}
\toprule
\multicolumn{4}{c}{Encoder Layers}
& Efficiency
& \multicolumn{2}{c}{LSVQ$_{\text{test}}$}
& \multicolumn{2}{c}{LSVQ$_{\text{1080p}}$}
& \multicolumn{2}{c}{KoNViD-1k}
& \multicolumn{2}{c}{LIVE-VQC}
& \multicolumn{2}{c}{Average} \\
\cmidrule(lr){1-4}\cmidrule(lr){5-5}\cmidrule(lr){6-7}\cmidrule(lr){8-9}\cmidrule(lr){10-11}\cmidrule(lr){12-13}\cmidrule(lr){14-15}
L3 & L6 & L9 & L12 & GFLOPs
& SROCC & PLCC & SROCC & PLCC & SROCC & PLCC & SROCC & PLCC & SROCC & PLCC \\
\midrule
           & \checkmark & \checkmark & \checkmark & 2403 & 0.905 & 0.905 & 0.830 & 0.861 & 0.897 & 0.900 & 0.856 & 0.878 & 0.872 & 0.886 \\
\checkmark &            & \checkmark & \checkmark & 2403 & 0.903 & 0.903 & 0.833 & 0.862 & 0.895 & 0.897 & 0.853 & 0.873 & 0.871 & 0.884 \\
\checkmark & \checkmark &            & \checkmark & 2403 & 0.903 & 0.903 & 0.830 & 0.860 & 0.894 & 0.896 & 0.860 & 0.879 & 0.872 & 0.885 \\
\checkmark & \checkmark & \checkmark &            & 2403 & 0.901 & 0.901 & 0.827 & 0.858 & 0.891 & 0.891 & 0.860 & 0.876 & 0.870 & 0.882 \\
\midrule
\checkmark &            &            &            & 2395 & 0.853 & 0.852 & 0.738 & 0.770 & 0.837 & 0.834 & 0.780 & 0.796 & 0.802 & 0.813 \\
           & \checkmark &            &            & 2395 & 0.889 & 0.889 & 0.798 & 0.835 & 0.881 & 0.879 & 0.845 & 0.862 & 0.853 & 0.867 \\
           &            & \checkmark &            & 2395 & 0.898 & 0.897 & 0.826 & 0.855 & 0.894 & 0.896 & 0.847 & 0.868 & 0.867 & 0.881 \\
           &            &            & \checkmark & 2395 & 0.896 & 0.896 & 0.828 & 0.855 & 0.891 & 0.894 & 0.846 & 0.871 & 0.865 & 0.879 \\
\midrule
\checkmark & \checkmark & \checkmark & \checkmark & 2406 & \textbf{0.906} & \textbf{0.906} & \textbf{0.835} & \textbf{0.867} & \textbf{0.902} & \textbf{0.903} & \textbf{0.864} & \textbf{0.882} & \textbf{0.877} & \textbf{0.890} \\
\bottomrule
\end{tabular}%
}
\end{table*}

\begin{table*}[t]
\centering
\caption{Ablation of the volumetric head and V-JEPA 2.1 encoder. ``Ridge Regressor''
replaces the volumetric network and token restructuring with a ridge regressor. ``ViT-B'' replaces the backbone with the
image-pretrained \texttt{vit\_base\_patch16\_224} checkpoint from
\texttt{timm} \cite{rw2019timm}, keeping the head fixed. We also compare with another unsupervised video backbone - VideoMAE \cite{tong2022videomae}.}
\label{tab:regressor_ablation}
\resizebox{1\linewidth}{!}{%
\begin{tabular}{l cc cc cc cc cc}
\toprule
\multirow{2}{*}{Quality Network (Backbone)}
& \multicolumn{2}{c}{LSVQ$_{\text{test}}$}
& \multicolumn{2}{c}{LSVQ$_{\text{1080p}}$}
& \multicolumn{2}{c}{KoNViD-1k}
& \multicolumn{2}{c}{LIVE-VQC}
& \multicolumn{2}{c}{Average} \\
\cmidrule(lr){2-3}\cmidrule(lr){4-5}\cmidrule(lr){6-7}\cmidrule(lr){8-9}\cmidrule(lr){10-11}
& SROCC & PLCC & SROCC & PLCC & SROCC & PLCC & SROCC & PLCC & SROCC & PLCC \\
\midrule
Ridge Regressor (V-JEPA 2.1 ViT-B) & 0.878 & 0.875 & 0.779 & 0.821 & 0.871 & 0.873 & 0.818 & 0.847 & 0.837 & 0.854 \\
\midrule
HiVE-VQA (ViT-B)       & 0.850 & 0.851 & 0.741 & 0.772 & 0.835 & 0.839 & 0.763 & 0.810 & 0.797 & 0.818 \\
HiVE-VQA (VideoMAE ViT-B) & 0.815 & 0.813 & 0.699 & 0.727 & 0.793 & 0.794 & 0.782 & 0.806 & 0.772 & 0.785 \\ 
\midrule
HiVE-VQA (V-JEPA 2.1 ViT-B)  & \textbf{0.906} & \textbf{0.906} & \textbf{0.835} & \textbf{0.867} & \textbf{0.902} & \textbf{0.903} & \textbf{0.864} & \textbf{0.882} & \textbf{0.877} & \textbf{0.890} \\
\bottomrule
\end{tabular}%
}
\end{table*}
\subsection{Encoder Representation Analysis}
\vspace{5pt}\noindent\textbf{Analysis of Embedding Sensitivity.} We visualize the representational capacity of our framework in Figure~\ref{fig:pca}. We take a video sequence, apply four different synthetic degradations, extract features from the encoder, and compute the first three principal components for each patch. These components are then mapped to RGB channels and plotted as an image. The resulting visualizations show that each type of distortion produces features with distinct representations, indicating that the V-JEPA 2.1 is indeed sensitive to various distortions.

\vspace{5pt}\noindent\textbf{Layer selection analysis.} To further understand the representational capabilities of the encoder, we evaluate every single-layer and leave-one-out configuration of the hierarchical volume (Table~\ref{tab:layer_ablation}). We first observe that hierarchical fusion offers an improvement ($+0.012$ SROCC) over the standard method of extracting features only from the final layer. This improvement is achieved at negligible additional computational cost (only 11 additional GFLOPs), as all layers come from a single forward pass through the frozen encoder. Further analysis suggests that no individual layer is indispensable to our framework, while the four-layer combination yields the most robust results, indicating that different layers may capture complementary information that generalizes differently across data distributions. Overall, the hierarchical fusion yields modest but consistent improvements, although the gain is smaller than that obtained from the feature volume, as shown in Table~\ref{tab:regressor_ablation}.

\subsection{Ablation Studies}
\vspace{5pt}\noindent\textbf{Volumetric Embedding Network Ablation.} We verify the effectiveness of the volumetric network by replacing it with a ridge regressor trained on the same multi-scale features and LSVQ training dataset (Table~\ref{tab:regressor_ablation}). The head improves average SROCC and PLCC by +0.040 and +0.036, respectively, confirming that a substantial gain in performance comes from the multi-scale volumetric network.

\vspace{5pt}\noindent\textbf{V-JEPA Backbone Ablation.} In order to understand the low-level representational capacity of JEPA features, we replace the V-JEPA 2.1 trained ViT-B encoder with one image pre-trained ViT-B \cite{rw2019timm} and VideoMAE ViT-B \cite{tong2022videomae}, which is another self-supervised video pre-training paradigm. Both of these encoders are ViT-B and have the same parameter count as the V-JEPA encoder. We train each replacement encoder under identical setups with the same HiVE-VQA framework. This ablation study shows that latent JEPA features greatly enhance performance with mean SROCC gains of +0.080 and +0.105, respectively. It is evident that features produced by JEPA pre-training hold powerful discriminative capacity for low-level representations.

\section{Conclusion}

In this paper, we are, to the best of our knowledge, the first to employ a JEPA-based encoder for NR-VQA, demonstrating its enhanced capability to represent low-level quality information. We propose a lightweight quality module that effectively captures both local and global context from a hierarchical feature volume. Extensive experiments demonstrate that the HiVE-VQA model achieves strong generalizability and accurate quality prediction across both cross-dataset and task-specific evaluations, with predictions that correlate well with human subjective judgments. Future work will focus on reducing the total parameter count through techniques such as knowledge distillation \cite{hinton2015distilling} and model pruning \cite{lecun1989optimal, zhu2021vision}.

\section{Acknowledgments}
The authors appreciate the funding from the EPSRC Doctoral Landscape Award, the University of Bristol, and the UKRI MyWorld Strength
in Places Program (SIPF00006/1). We would like to thank Benoit Vallade and Alex Mackin from Amazon Prime
Video, UK, for their valuable advice and support throughout this work. We also acknowledge the use of resources provided by the Isambard-AI National AI Research Resource (AIRR). Isambard-AI is operated by the University of Bristol and is funded by the UK Government’s Department for Science, Innovation and Technology (DSIT) via UK Research and Innovation; and the Science and Technology Facilities Council
[ST/AIRR/I-A-I/1023].
{
    \small
    \bibliographystyle{ieeenat_fullname}
    \bibliography{main}

@String(CVPR= {IEEE Conf. Comput. Vis. Pattern Recog.})

@String(ICCV= {Int. Conf. Comput. Vis.})

@String(ECCV= {Eur. Conf. Comput. Vis.})

@String(ICASSP=	{ICASSP})

@String(CVPR  = {CVPR})

@String(ICCV  = {ICCV})

@String(ECCV  = {ECCV})

@misc{cisco2020air,
  author = {{Cisco}},
  title = {Cisco Annual Internet Report (2018--2023) White Paper},
  year = {2020},
  month = mar,
  howpublished = {\url{https://www.cisco.com/c/en/us/solutions/collateral/executive-perspectives/annual-internet-report/white-paper-c11-741490.html}}
}

@ARTICLE{bliinds,
  author={Saad, Michele A. and Bovik, Alan C. and Charrier, Christophe},
  journal={IEEE Transactions on Image Processing}, 
  title={Blind Prediction of Natural Video Quality}, 
  year={2014},
  volume={23},
  number={3},
  pages={1352-1365},
  doi={10.1109/TIP.2014.2299154}}

@article{brisque,
  title={No-reference image quality assessment in the spatial domain},
  author={Mittal, Anish and Moorthy, Anush Krishna and Bovik, Alan Conrad},
  journal={IEEE Transactions on image processing},
  volume={21},
  number={12},
  pages={4695--4708},
  year={2012},
  publisher={IEEE}
}

@ARTICLE{viideo,
  author={Mittal, Anish and Saad, Michele A. and Bovik, Alan C.},
  journal={IEEE Transactions on Image Processing}, 
  title={A Completely Blind Video Integrity Oracle}, 
  year={2016},
  volume={25},
  number={1},
  pages={289-300},
  doi={10.1109/TIP.2015.2502725}}

@ARTICLE{chipqa,
  author={Ebenezer, Joshua Peter and Shang, Zaixi and Wu, Yongjun and Wei, Hai and Sethuraman, Sriram and Bovik, Alan C.},
  journal={IEEE Transactions on Image Processing}, 
  title={ChipQA: No-Reference Video Quality Prediction via Space-Time Chips}, 
  year={2021},
  volume={30},
  number={},
  pages={8059-8074},
  doi={10.1109/TIP.2021.3112055}}

@ARTICLE{niqe,
  author={Mittal, Anish and Soundararajan, Rajiv and Bovik, Alan C.},
  journal={IEEE Signal Processing Letters}, 
  title={Making a “Completely Blind” Image Quality Analyzer}, 
  year={2013},
  volume={20},
  number={3},
  pages={209-212},
  doi={10.1109/LSP.2012.2227726}}

@ARTICLE{tlvqm,
  author={Korhonen, Jari},
  journal={IEEE Transactions on Image Processing}, 
  title={Two-Level Approach for No-Reference Consumer Video Quality Assessment}, 
  year={2019},
  volume={28},
  number={12},
  pages={5923-5938},
  doi={10.1109/TIP.2019.2923051}}

@inproceedings{cnntlvqm,
author = {Korhonen, Jari and Su, Yicheng and You, Junyong},
title = {Blind Natural Video Quality Prediction via Statistical Temporal Features and Deep Spatial Features},
year = {2020},
isbn = {9781450379885},
publisher = {Association for Computing Machinery},
address = {New York, NY, USA},
url = {https://doi.org/10.1145/3394171.3413845},
doi = {10.1145/3394171.3413845},
booktitle = {Proceedings of the 28th ACM International Conference on Multimedia},
pages = {3311–3319},
numpages = {9},
location = {Seattle, WA, USA},
series = {MM '20}
}

@ARTICLE{videval,
  author={Tu, Zhengzhong and Wang, Yilin and Birkbeck, Neil and Adsumilli, Balu and Bovik, Alan C.},
  journal={IEEE Transactions on Image Processing}, 
  title={UGC-VQA: Benchmarking Blind Video Quality Assessment for User Generated Content}, 
  year={2021},
  volume={30},
  number={},
  pages={4449-4464},
  doi={10.1109/TIP.2021.3072221}}

@InProceedings{patchvq,
    author    = {Ying, Zhenqiang and Mandal, Maniratnam and Ghadiyaram, Deepti and Bovik, Alan},
    title     = {Patch-VQ: 'Patching Up' the Video Quality Problem},
    booktitle = {Proceedings of the IEEE/CVF Conference on Computer Vision and Pattern Recognition (CVPR)},
    month     = {June},
    year      = {2021},
    pages     = {14019-14029}
}

@ARTICLE{bvqa,
  author={Li, Bowen and Zhang, Weixia and Tian, Meng and Zhai, Guangtao and Wang, Xianpei},
  journal={IEEE Transactions on Circuits and Systems for Video Technology}, 
  title={Blindly Assess Quality of In-the-Wild Videos via Quality-Aware Pre-Training and Motion Perception}, 
  year={2022},
  volume={32},
  number={9},
  pages={5944-5958},
  doi={10.1109/TCSVT.2022.3164467}}

@inproceedings{dover,
      title={Exploring Video Quality Assessment on User Generated Contents from Aesthetic and Technical Perspectives}, 
      author={Wu, Haoning and Zhang, Erli and Liao, Liang and Chen, Chaofeng and Hou, Jingwen and Wang, Annan and Sun, Wenxiu and Yan, Qiong and Lin, Weisi},
      year={2023},
      booktitle={International Conference on Computer Vision (ICCV)},
}

@inproceedings{fast-vqa,
  title={FAST-VQA: Efficient End-to-end Video Quality Assessment with Fragment Sampling},
  author={Wu, Haoning and Chen, Chaofeng and Hou, Jingwen and Liao, Liang and Wang, Annan and Sun, Wenxiu and Yan, Qiong and Lin, Weisi},
  booktitle ={Proceedings of European Conference of Computer Vision (ECCV)},
  year={2022}
}

@InProceedings{zoomvqa,
    author    = {Zhao, Kai and Yuan, Kun and Sun, Ming and Wen, Xing},
    title     = {Zoom-VQA: Patches, Frames and Clips Integration for Video Quality Assessment},
    booktitle = {Proceedings of the IEEE/CVF Conference on Computer Vision and Pattern Recognition (CVPR) Workshops},
    month     = {June},
    year      = {2023},
    pages     = {1302-1310}
}

@article{qalign,
  title={Q-Align: Teaching LMMs for Visual Scoring via Discrete Text-Defined Levels},
  author={Wu, Haoning and Zhang, Zicheng and Zhang, Weixia and Chen, Chaofeng and Li, Chunyi and Liao, Liang and Wang, Annan and Zhang, Erli and Sun, Wenxiu and Yan, Qiong and Min, Xiongkuo and Zhai, Guangtao and Lin, Weisi},
  journal={arXiv preprint arXiv:2312.17090},
  year={2023},
  institution={Nanyang Technological University and Shanghai Jiao Tong University and Sensetime Research},
}

@inproceedings{clifvqa,
author = {Mi, Yachun and Shu, Yan and Li, Yu and Hui, Chen and Zhou, Puchao and Liu, Shaohui},
title = {CLiF-VQA: Enhancing Video Quality Assessment by Incorporating High-Level Semantic Information related to Human Feelings},
year = {2024},
isbn = {9798400706868},
publisher = {Association for Computing Machinery},
address = {New York, NY, USA},
url = {https://doi.org/10.1145/3664647.3680930},
doi = {10.1145/3664647.3680930},
booktitle = {Proceedings of the 32nd ACM International Conference on Multimedia},
pages = {9989–9998},
numpages = {10},
location = {Melbourne VIC, Australia},
series = {MM '24}
}

@InProceedings{mvqa,
    author    = {Mi, Yachun and Li, Yu and Meng, Weicheng and Chen, Chaofeng and Hui, Chen and Liu, Shaohui},
    title     = {MVQA: Mamba with Unified Sampling for Efficient Video Quality Assessment},
    booktitle = {Proceedings of the IEEE/CVF International Conference on Computer Vision (ICCV)},
    month     = {October},
    year      = {2025},
    pages     = {18498-18509}
}

@inproceedings{vsfa,
  title={Quality assessment of in-the-wild videos},
  author={Li, Dingquan and Jiang, Tingting and Jiang, Ming},
  booktitle={Proceedings of the 27th ACM international conference on multimedia},
  pages={2351--2359},
  year={2019}
}

@article{faster-vqa,
  title={Neighbourhood representative sampling for efficient end-to-end video quality assessment},
  author={Wu, Haoning and Chen, Chaofeng and Liao, Liang and Hou, Jingwen and Sun, Wenxiu and Yan, Qiong and Gu, Jinwei and Lin, Weisi},
  journal={IEEE Transactions on Pattern Analysis and Machine Intelligence},
  volume={45},
  number={12},
  pages={15185--15202},
  year={2023},
  publisher={IEEE}
}

@InProceedings{mbvqa,
    author    = {Wen, Wen and Li, Mu and Zhang, Yabin and Liao, Yiting and Li, Junlin and Zhang, Li and Ma, Kede},
    title     = {Modular Blind Video Quality Assessment},
    booktitle = {Proceedings of the IEEE/CVF Conference on Computer Vision and Pattern Recognition (CVPR)},
    month     = {June},
    year      = {2024},
    pages     = {2763-2772}
}

@ARTICLE{rapique,
  author={Tu, Zhengzhong and Yu, Xiangxu and Wang, Yilin and Birkbeck, Neil and Adsumilli, Balu and Bovik, Alan C.},
  journal={IEEE Open Journal of Signal Processing}, 
  title={RAPIQUE: Rapid and Accurate Video Quality Prediction of User Generated Content}, 
  year={2021},
  volume={2},
  number={},
  pages={425-440},
  doi={10.1109/OJSP.2021.3090333}}

@InProceedings{coinvq,
    author    = {Wang, Yilin and Ke, Junjie and Talebi, Hossein and Yim, Joong Gon and Birkbeck, Neil and Adsumilli, Balu and Milanfar, Peyman and Yang, Feng},
    title     = {Rich Features for Perceptual Quality Assessment of UGC Videos},
    booktitle = {Proceedings of the IEEE/CVF Conference on Computer Vision and Pattern Recognition (CVPR)},
    month     = {June},
    year      = {2021},
    pages     = {13435-13444}
}

@inproceedings{maxvqa,
author = {Wu, Haoning and Zhang, Erli and Liao, Liang and Chen, Chaofeng and Hou, Jingwen and Wang, Annan and Sun, Wenxiu and Yan, Qiong and Lin, Weisi},
title = {Towards Explainable In-the-Wild Video Quality Assessment: A Database and a Language-Prompted Approach},
year = {2023},
isbn = {9798400701085},
publisher = {Association for Computing Machinery},
address = {New York, NY, USA},
url = {https://doi.org/10.1145/3581783.3611737},
doi = {10.1145/3581783.3611737},
booktitle = {Proceedings of the 31st ACM International Conference on Multimedia},
pages = {1045–1054},
numpages = {10},
location = {Ottawa ON, Canada},
series = {MM '23}
}

@ARTICLE{gst-vqa,
  author={Chen, Baoliang and Zhu, Lingyu and Li, Guo and Lu, Fangbo and Fan, Hongfei and Wang, Shiqi},
  journal={IEEE Transactions on Circuits and Systems for Video Technology}, 
  title={Learning Generalized Spatial-Temporal Deep Feature Representation for No-Reference Video Quality Assessment}, 
  year={2022},
  volume={32},
  number={4},
  pages={1903-1916},
  doi={10.1109/TCSVT.2021.3088505}}

@misc{konvid1k,
title = {The Konstanz Natural Video Database},
author = {Hosu, Vlad and Hahn, Franz and Jenadeleh, 
Mohsen and Lin, Hanhe and Men, Hui and Szir{\'a}nyi,
Tam{\'a}s and Li, Shujun and Saupe, Dietmar},
year = {2017},
url = {http://database.mmsp-kn.de}}

@inproceedings{hosu2017konstanz,
title = {The Konstanz natural video database 
(KoNViD-1k)},
author = {Hosu, Vlad and Hahn, Franz and Jenadeleh, 
Mohsen and Lin, Hanhe and Men, Hui and Szir{\'a}nyi,
Tam{\'a}s and Li, Shujun and Saupe, Dietmar},
booktitle = {2017 Ninth International Conference on 
Quality of Multimedia Experience (QoMEX)},
pages = {1--6},
year = {2017},
organization = {IEEE}}

@ARTICLE{livevqc,
  author={Sinno, Zeina and Bovik, Alan Conrad},
  journal={IEEE Transactions on Image Processing}, 
  title={Large-Scale Study of Perceptual Video Quality}, 
  year={2019},
  volume={28},
  number={2},
  pages={612-627},
  doi={10.1109/TIP.2018.2869673}}

@INPROCEEDINGS{youtubeugc,
  author={Wang, Yilin and Inguva, Sasi and Adsumilli, Balu},
  booktitle={2019 IEEE 21st International Workshop on Multimedia Signal Processing (MMSP)}, 
  title={YouTube UGC Dataset for Video Compression Research}, 
  year={2019},
  volume={},
  number={},
  pages={1-5},
  doi={10.1109/MMSP.2019.8901772}}

@InProceedings{jepa,
    author    = {Assran, Mahmoud and Duval, Quentin and Misra, Ishan and Bojanowski, Piotr and Vincent, Pascal and Rabbat, Michael and LeCun, Yann and Ballas, Nicolas},
    title     = {Self-Supervised Learning From Images With a Joint-Embedding Predictive Architecture},
    booktitle = {Proceedings of the IEEE/CVF Conference on Computer Vision and Pattern Recognition (CVPR)},
    month     = {June},
    year      = {2023},
    pages     = {15619-15629}
}

@article{vjepa2,
  title={V-JEPA~2: Self-Supervised Video Models Enable Understanding, Prediction and Planning},
  author={Assran, Mahmoud and Bardes, Adrien and Fan, David and Garrido, Quentin and Howes, Russell and
Komeili, Mojtaba and Muckley, Matthew and Rizvi, Ammar and Roberts, Claire and Sinha, Koustuv and Zholus, Artem and
Arnaud, Sergio and Gejji, Abha and Martin, Ada and Robert Hogan, Francois and Dugas, Daniel and
Bojanowski, Piotr and Khalidov, Vasil and Labatut, Patrick and Massa, Francisco and Szafraniec, Marc and
Krishnakumar, Kapil and Li, Yong and Ma, Xiaodong and Chandar, Sarath and Meier, Franziska and LeCun, Yann and
Rabbat, Michael and Ballas, Nicolas},
  journal={arXiv preprint arXiv:2506.09985},
  year={2025}
}

@article{vjepa2_1,
  title={V-JEPA 2.1: Unlocking Dense Features in Video Self-Supervised Learning},
  author={Mur-Labadia, Lorenzo and Muckley, Matthew and Bar, Amir and Assran, Mahmoud and
Sinha, Koustuv and Rabbat, Michael and LeCun, Yann and Ballas, Nicolas and Bardes, Adrien},
  journal={arXiv preprint arXiv:2603.14482},
  year={2026}
}

@ARTICLE{contrique,
  author={Madhusudana, Pavan C. and Birkbeck, Neil and Wang, Yilin and Adsumilli, Balu and Bovik, Alan C.},
  journal={IEEE Transactions on Image Processing}, 
  title={Image Quality Assessment Using Contrastive Learning}, 
  year={2022},
  volume={31},
  number={},
  pages={4149-4161},
  doi={10.1109/TIP.2022.3181496}}

@ARTICLE{conviqt,
  author={Madhusudana, Pavan C. and Birkbeck, Neil and Wang, Yilin and Adsumilli, Balu and Bovik, Alan C.},
  journal={IEEE Transactions on Image Processing}, 
  title={CONVIQT: Contrastive Video Quality Estimator}, 
  year={2023},
  volume={32},
  number={},
  pages={5138-5152},
  doi={10.1109/TIP.2023.3310344}}

@InProceedings{rankdvqa,
    author    = {Feng, Chen and Danier, Duolikun and Zhang, Fan and Bull, David},
    title     = {RankDVQA: Deep VQA Based on Ranking-Inspired Hybrid Training},
    booktitle = {Proceedings of the IEEE/CVF Winter Conference on Applications of Computer Vision (WACV)},
    month     = {January},
    year      = {2024},
    pages     = {1648-1658}
}

@INPROCEEDINGS{rankdvqamini,
  author={Feng, Chen and Danier, Duolikun and Wang, Haoran and Zhang, Fan and Vallade, Benoit and Mackin, Alex and Bull, David},
  booktitle={2024 Picture Coding Symposium (PCS)}, 
  title={RankDVQA-Mini: Knowledge Distillation-Driven Deep Video Quality Assessment}, 
  year={2024},
  volume={},
  number={},
  pages={1-5},
  doi={10.1109/PCS60826.2024.10566364}}

@ARTICLE{dvlta-vqa,
  author={Yu, Li and Wang, Situo and Zhou, Wei and Gabbouj, Moncef},
  journal={IEEE Transactions on Circuits and Systems for Video Technology}, 
  title={DVLTA-VQA: Decoupled Vision-Language Modeling With Text-Guided Adaptation for Blind Video Quality Assessment}, 
  year={2026},
  volume={36},
  number={5},
  pages={6826-6837},
  doi={10.1109/TCSVT.2026.3657415}}

@article{hinton2015distilling,
  title={Distilling the knowledge in a neural network},
  author={Hinton, Geoffrey and Vinyals, Oriol and Dean, Jeff},
  journal={arXiv preprint arXiv:1503.02531},
  year={2015}
}

@article{lecun1989optimal,
  title={Optimal brain damage},
  author={LeCun, Yann and Denker, John and Solla, Sara},
  journal={Advances in neural information processing systems},
  volume={2},
  year={1989}
}

@article{zhu2021vision,
  title={Vision transformer pruning},
  author={Zhu, Mingjian and Tang, Yehui and Han, Kai},
  journal={arXiv preprint arXiv:2104.08500},
  year={2021}
}

@techreport{vqeg2000,
author = {{Video Quality Experts Group}},
title = {Final Report from the Video Quality Experts Group on the Validation of Objective Quality Metrics for Video Quality Assessment},
year = {2000},
institution = {Video Quality Experts Group},
type = {Technical Report}
}

@online{vmaf,
  author  = {Zhi Li and Anne Aaron and Ioannis Katsavounidis and Anush Moorthy and Megha Manohara},
  title   = {Toward A Practical Perceptual Video Quality Metric},
  year    = {2016},
  month   = jun,
  day     = {6},
  url     = {https://netflixtechblog.com/toward-a-practical-perceptual-video-quality-metric-653f208b9652},
  urldate = {2026-08-10},
  note    = {Netflix Technology Blog}
}

@inproceedings{li2020norm,
  title={Norm-in-norm loss with faster convergence and better performance for image quality assessment},
  author={Li, Dingquan and Jiang, Tingting and Jiang, Ming},
  booktitle={Proceedings of the 28th ACM International conference on multimedia},
  pages={789--797},
  year={2020}
}

@article{li_unified_2021,
	title = {Unified {Quality} {Assessment} of in-the-{Wild} {Videos} with {Mixed} {Datasets} {Training}},
	volume = {129},
	issn = {0920-5691, 1573-1405},
	url = {http://link.springer.com/10.1007/s11263-020-01408-w},
	doi = {10.1007/s11263-020-01408-w},
	language = {en},
	number = {4},
	urldate = {2026-08-11},
	journal = {Int J Comput Vis},
	author = {Li, Dingquan and Jiang, Tingting and Jiang, Ming},
	month = apr,
	year = {2021},
	pages = {1238--1257},
}

@inproceedings{ma2018geometric,
  title={Geometric transformation invariant image quality assessment using convolutional neural networks},
  author={Ma, Kede and Duanmu, Zhengfang and Wang, Zhou},
  booktitle={2018 IEEE International Conference on Acoustics, Speech and Signal Processing (ICASSP)},
  pages={6732--6736},
  year={2018},
  organization={IEEE}
}

@inproceedings{liu2018end,
  title={End-to-end blind quality assessment of compressed videos using deep neural networks.},
  author={Liu, Wentao and Duanmu, Zhengfang and Wang, Zhou},
  booktitle={ACM Multimedia},
  volume={39},
  pages={546--554},
  year={2018}
}

@article{li2021user,
  title={User-generated video quality assessment: A subjective and objective study},
  author={Li, Yang and Meng, Shengbin and Zhang, Xinfeng and Wang, Meng and Wang, Shiqi and Wang, Yue and Ma, Siwei},
  journal={IEEE Transactions on Multimedia},
  volume={25},
  pages={154--166},
  year={2021},
  publisher={IEEE}
}

@inproceedings{qu2025kvq,
  title={KVQ: boosting video quality assessment via saliency-guided local perception},
  author={Qu, Yunpeng and Yuan, Kun and Xie, Qizhi and Sun, Ming and Zhou, Chao and Wang, Jian},
  booktitle={2025 IEEE/CVF Conference on Computer Vision and Pattern Recognition (CVPR)},
  pages={2150--2160},
  year={2025},
  organization={IEEE}
}

@article{vaswani2017attention,
  title={Attention is all you need},
  author={Vaswani, Ashish and Shazeer, Noam and Parmar, Niki and Uszkoreit, Jakob and Jones, Llion and Gomez, Aidan N and Kaiser, {\L}ukasz and Polosukhin, Illia},
  journal={Advances in neural information processing systems},
  volume={30},
  year={2017}
}

@article{dosovitskiy2020image,
  title={An image is worth 16x16 words: Transformers for image recognition at scale},
  author={Dosovitskiy, Alexey and Beyer, Lucas and Kolesnikov, Alexander and Weissenborn, Dirk and Zhai, Xiaohua and Unterthiner, Thomas and Dehghani, Mostafa and Minderer, Matthias and Heigold, Georg and Gelly, Sylvain and others},
  journal={arXiv preprint arXiv:2010.11929},
  year={2020}
}

@article{tong2022videomae,
  title={Videomae: Masked autoencoders are data-efficient learners for self-supervised video pre-training},
  author={Tong, Zhan and Song, Yibing and Wang, Jue and Wang, Limin},
  journal={Advances in neural information processing systems},
  volume={35},
  pages={10078--10093},
  year={2022}
}

@misc{rw2019timm,
  author = {Ross Wightman},
  title = {PyTorch Image Models},
  year = {2019},
  publisher = {GitHub},
  journal = {GitHub repository},
  doi = {10.5281/zenodo.4414861},
  howpublished = {\url{https://github.com/rwightman/pytorch-image-models}}
}

@article{pearson1895note,
  title={Note on regression and inheritance in the case of two parents},
  author={Pearson, Karl},
  journal={proceedings of the royal society of London},
  volume={58},
  pages={240--242},
  year={1895},
  publisher={JSTOR}
}

@article{spearman1904proof,
  author  = {Spearman, Charles},
  title   = {The Proof and Measurement of Association between Two Things},
  journal = {The American Journal of Psychology},
  volume  = {15},
  pages   = {72--101},
  year    = {1904}
}
}

\end{document}